\documentclass[10pt,twocolumn,letterpaper]{article}
\usepackage{authblk}
\usepackage{wacv}
\usepackage{times}
\usepackage{epsfig}
\usepackage{graphicx}
\usepackage{amsmath}
\usepackage{amssymb}
\usepackage[cjk]{kotex}
\wacvfinalcopy % *** Uncomment this line for the final submission

\ifwacvfinal
\fi

\begin{document}

%%%%%%%%% TITLE
\title{Painting Outside as Inside:  Edge Guided Image Outpainting via Bidirectional Rearrangement with Progressive Step Learning}

% \author{Kyunghun Kim, Yeohun Yun, Keon-Woo Kang\\
% Department of Electronic Engineering, Sogang University, \\
%  Seoul, South Korea\\
% {\tt\small godgang@sogang.ac.kr}
% \and
% Siyeong Lee\\
% NaverLABs
% % Department of Electronic Engineering, Sogang University, \\
% %  Seoul, South Korea\\
% {\tt\small dugns159@gmail.com}
% {\tt\small gunkun00@gmail.com}
% {\tt\small siyeong.lee@naverlabs.com}
% {\tt\small kkb4723@postech.ac.kr}
% }

\author[1]{Kyunghun Kim}
\author[1]{Yeohun Yun}
\author[1]{Keon-Woo Kang}
\author[2]{Kyeongbo Kong}
\author[3]{Siyeong Lee}
\author[1]{Suk-Ju Kang}
\affil[1]{Department of Electronic Engineering, Sogang University \\
  Seoul, South Korea,\\}
\affil[2]{Department of Electrical Engineering, POSTECH, Pohang, South Korea,}
\affil[3]{NAVER LABS, South Korea}
\affil[ ]{\textit {godgang@sogang.ac.kr, dugns159@gmail.com, gunkun00@gmail.com\\
kkb4723@postech.ac.kr, siyeong.lee@naverlabs.com, sjkang@sogang.ac.kr}}

\maketitle
%\thispagestyle{empty}

%%%%%%%%% ABSTRACT
\begin{abstract}
   Image outpainting is a very intriguing problem as the outside of a given image can be continuously filled by considering as the context of the image. This task has two main challenges. The first is to maintain the spatial consistency in contents of generated regions and the original input. The second is to generate a high-quality large image with a small amount of adjacent information. Conventional image outpainting methods generate inconsistent, blurry, and repeated pixels. To alleviate the difficulty of an outpainting problem, we propose a novel image outpainting method using bidirectional boundary region rearrangement. We rearrange the image to benefit from the image inpainting task by reflecting more directional information. The bidirectional boundary region rearrangement enables the generation of the missing region using bidirectional information similar to that of the image inpainting task, thereby generating the higher quality than the conventional methods using unidirectional information. Moreover, we use the edge map generator that considers images as original input with structural information and hallucinates  the edges of unknown regions to generate the image. Our proposed method is compared with other state-of-the-art outpainting and inpainting methods both qualitatively and quantitatively. We further compared and evaluated them using BRISQUE, one of the No-Reference image quality assessment (IQA) metrics, to evaluate the naturalness of the output. The experimental results demonstrate that our method outperforms other methods and generates new images with 360°panoramic characteristics.
\end{abstract}

%%%%%%%%% BODY TEXT
\begin{figure}
\centering
\includegraphics[width=80mm]{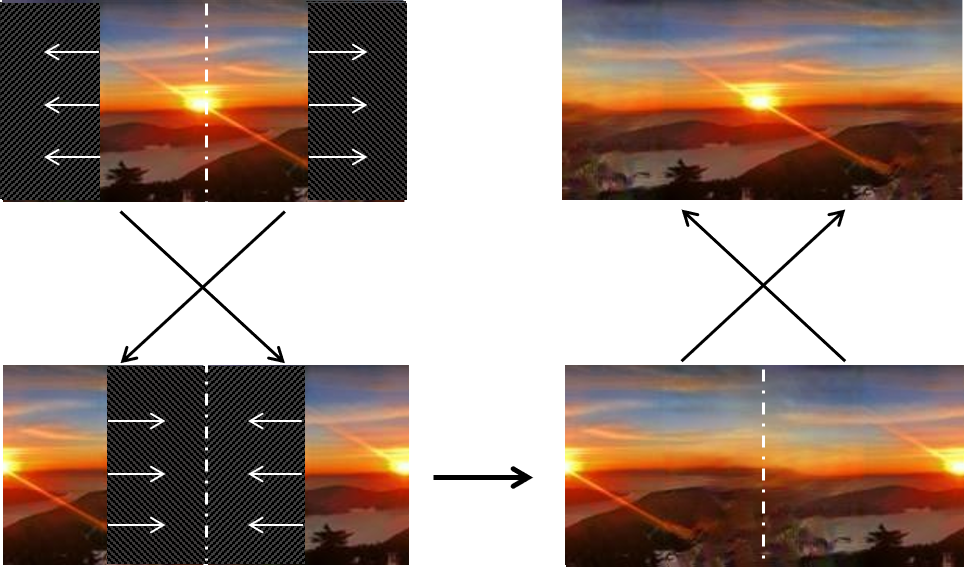} 
\caption{
Main concept of the proposed method. We rearrange the outer area of the image into its inner area. This enables us to consider bidirectional information, and it can generate a high-quality natural image that is superior to images produced by conventional methods.}
\label{fig:1}
\end{figure}

% \begin{figure*}
% \centering
% \includegraphics[width=\textwidth]{fig1_3.png} 
% \caption{
% Panoramic image result through our model based on multi-steps with internal and external extrapolation.}
% \label{fig:1}
% \end{figure*}
% \section{Introduction}

\section{Introduction}

With developments in deep learning, image completion has been actively researched and it has had the great performance for various applications. Typically, image completion includes image inpainting \cite{nazeri2019edgeconnect,pathak2016context,yu2018generative} and outpainting \cite{iizuka2017globally,sabini2018painting,van2019image}. Image inpainting predicts relatively small missing or corrupted parts of the interiors in a given image based on the surrounding pixels. Image outpainting is used to extend images beyond their borders, thereby generating larger images with added peripheries. The task can be classified into two categories according to the amount of extrapolation desired: slightly extending images can be useful in such applications as panorama construction, when small additions may be needed to make the panorama rectangular after image stitching \cite{wang2014biggerpicture}. Classical image outpainting methods are patch-based methods \cite{kopf2012quality,sivic2008creating,zhang2013framebreak} that fill in the missing areas by copying similar information from the known areas of the image. These methods \cite{kopf2012quality,sivic2008creating,zhang2013framebreak} perform well when filling small missing areas on a simple texture image, but the results are semantically unnatural and inconsistent with the given context. In recent years, generative adversarial network (GAN)-based approaches   \cite{goodfellow2014generative} have addressed the limitations of classical methods. In particular, the GAN network used in image outpainting that generates a large region is exceedingly difficult to train with stability. 

Image inpainting and outpainting methods tend to perform a similar task with regard to filling the missing regions or unknown regions of a given image. Although image outpainting is closely related to image inpainting, which has been making considerable progress recently, image outpainting has attracted less attention because it is more difficult compared to inpainting. First, image inpainting fills the missing region using adjacent and bidirectional information; however, image outpainting fills the missing region using only unidirectional information. As image outpainting can utilize less information than image inpainting, image outpainting methods generate poor-quality output images as compared to those obtained by inpainting. Second, image outpainting generates a larger unknown region than that in image inpainting; hence, maintaining spatial consistency of contents between an input image and generated regions is difficult.

Existing image outpainting methods \cite{van2019image,yang2019very} using GANs still use only one side of adjacent information, resulting in blurry textures. To solve image outpainting problems, how about restoring the image as if it were a panoramic image? We propose a bidirectional boundary region rearrangement method to increase the adjacent information. The overall of the rearrangement method shown  in  Fig.~\ref{fig:1}. We rearrange the outer area of the image into its inner area. This enables us to consider bidirectional information, and it can generate a high-quality natural image that is superior to images produced by conventional methods. In addition, we use the structural edge map generator to make the structural information clearer. Moreover, we present a progressive step learning method that divides the learning step into multiple steps to stabilize the GAN network. The method is based on the gains coarse-to-fine learning strategy  \cite{abbas2019learning} used in image inpainting. For more efficient outpainting task, we converted the progressive learning method \cite{abbas2019learning} into a method that increases only the horizontal area of the mask to better connect the information at both ends of the horizontal area. We trained the model by step by step increasing the size of the mask per step. 

In summary, our contributions are as follows.

 1. We propose a novel bidirectional boundary region rearrangement method to alleviate the difficulty of a problem by changing the problem domain from image outpainting to image inpainting. Using this approach, our proposed method can generate a semantically more natural image better than conventional methods. Not only can we extend the image outside we also can extend the image inside the image. 
 
2. We present a hinge loss specialized in an edge map generator. We use a structural edge map generator optimized for image outpainting. It generates an edge map to make the structural information clearer, thereby generating a photo-realistic image considering the surrounding regional characteristics. 
% We applied the Swish[] activation function and confirmed the performance improvement.

3. We introduce a horizontally progressive step learning method that stabilizes the GAN generator and better connects the information in the horizontal region at both ends. This method is useful for horizontal image outpainting that generates a large unknown region. This can also achieve an augmentation effect for small datasets.

\begin{figure*}[t!]
\centering
\includegraphics[width=\textwidth]{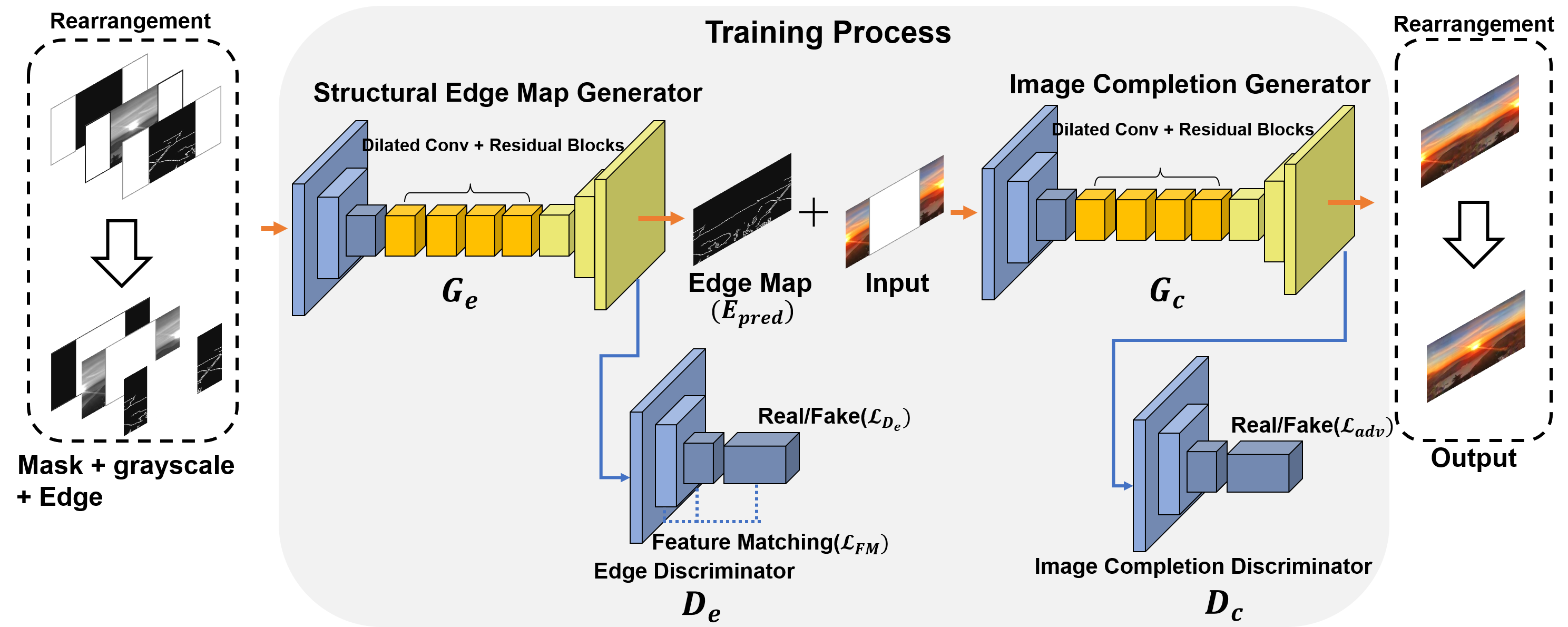}
\caption{Overall architecture of our image outpainting network. Our model comprises two parts: an edge map generation network and an image completion network. Each part is composed of a generator-discriminator pair.}
\label{fig:2}
\end{figure*}
%-------------------------------------------------------------------------
\section{Related Work}
\textbf{Image Inpainting} This method fills the unknown region in input image, and then make the image photo-realistic by extracting the image features of the image through the information of the known region. Conventional image inpainting methods \cite{barnes2009patchmatch,efros2001image} rely on the similarity or diffusion of a patch to obtain information regarding unknown regions from known regions. These methods are effective when the damaged region is small. However, if the unknown region becomes large, they cannot perform semantic analysis and consequently generate image with low quality. The use of deep learning-based generative adversarial networks (GANs) \cite{goodfellow2014generative} for image generation led to improved performance of image inpainting methods. Iizuka et al. \cite{iizuka2017globally} proposed a method using two discriminators—a global discriminator and a local discriminator—based on GANs. The global discriminator scans the entire image to assess its coherency, whereas the local discriminator scans only a small area centered at the completed region to ensure the local consistency of the generated patches. This allows them to generate a naturally unfragmented image. Liu et al. \cite{liu2018partial} proposed partial convolution that is masked and renormalized to be conditioned only on valid pixels. Typically, image inpainting methods use a standard convolutional network over the corrupted image, using convolutional filters conditioned on both valid pixels as well as pixels in the unknown regions. This often leads to artifacts such as color discrepancy and blurriness. However, as the partial convolution \cite{liu2018partial} renormalized convolution operation is conditioned only on valid pixels, it can robustly fill the missing regions of any shape, size, location, or distance from the image borders. Yu et al. \cite{yu2018generative} proposed a two-stage coarse-to-fine network architecture, in which the first network makes an initial coarse prediction, and the second network takes the coarse prediction as inputs and predicts the refined results. In addition, Yu et al. \cite{yu2018generative} proposed a contextual attention module that copied significant features from the known region to the unknown region that preserved the details of the filled region. Inpainting algorithms tend to have more predictable and higher-quality results compared to those of outpainting algorithms. However, we here in demonstrate that using inpainting algorithms with no modifications leads to poor results for image outpainting.

\textbf{Image Outpainting} Image outpainting is a method that naturally fills the external region of the image to generate a natural image similar to that obtained by image inpainting. However, as mentioned in the previous section, the area to be generated is typically larger than that in image inpainting, and the adjacent information is insufficient. Conventional image outpainting methods \cite{kopf2012quality,sivic2008creating,zhang2013framebreak} select candidates through the similarity of a patch, but their performance becomes less reliable as the size of the generated region increases. Recently, some methods \cite{gatys2016image,johnson2016perceptual} using GANs have been proposed to enhance the output image quality. Sabini el al. \cite{sabini2018painting} proposed a method based on GANs for the first time. This method comprises a simple encoder–decoder structure and uses only the mean squared error (MSE) and GAN loss \cite{goodfellow2014generative}. It uses postprocessing to smooth the output image but shows repeated pixels in the generated image. Teterwak et al. \cite{teterwak2019boundless} used semantic conditioning in the discriminator, which is a stabilization scheme for training, based on the semantic information from a pre-trained deep network to regulate the behavior of the discriminator. Yang et al. \cite{yang2019very} proposed a recurrent content transfer (RCT) model based on long short-term memory (LSTM) \cite{hochreiter1997long}. RCT transfers the input region features to the prediction region, improving a natural connection between the input region and the prediction region. However, the generated region reflects the features of the adjacent input region relatively largely; thus, the entire generated region tends to not match the color tone, which is an important feature of the input image. To solve this problem, we adapt two losses proposed in \cite{gatys2016image,johnson2016perceptual}, known as style loss and perceptual loss. Using these losses, the semantic contents are RAT consistent in styles such as color tones of the input image.
\section{Methodology}
We provide an overview of the overall architecture, which is shown in Fig.~\ref{fig:2}, then provide details on each component. Our model comprises two parts: an edge map generation network and an image completion network. Each part is composed of a generator-discriminator pair. We define the generator and discriminator of the structural edge map generator as $G_e$ and $D_e$, respectively. $G_e$ is used to predict missing structures, thereby generating the global edge map image, $E_{pred}$, which is used in the image completion network. The generator and discriminator of the image completion network are $G_c$ and $D_c$, respectively. $G_c$ draws details according to the edge map image and generates the final image, $I_{pred}$. Our generators follow an architecture similar to that proposed by Johnson et al. \cite{johnson2016perceptual} and uses instance normalization \cite{ulyanov2017improved} across all layers of the network.

\subsection{Bidirectional Boundary Region Rearrangement}
Image outpainting, which only utilizes unidirectional information, is less reliable than image inpainting, which generates missing areas through bidirectional information. Previous methods  \cite{sabini2018painting,yang2019very,teterwak2019boundless} use only unidirectional information and generate structurally and semantically insufficient images. To overcome this difficulty in image outpainting, we rearranged the image to benefit from the image inpainting by considering more directional information. We  propose  a  novel  bidirectional  boundary  region rearrangement to alleviate the difficulty of a problem by changing the problem domain from image outpainting to image  inpainting. When testing, our model preprocesses an input image through  bidirectional boundary region rearrangement and performs the same network process as training. The final image also rearranges the output image through the rearrangement module once again.

\subsection{Structural Edge Map Generator }
An edge map is usually used to enforce structural quality prior to image inpainting \cite{nazeri2019edgeconnect} and image super-resolution \cite{nazeri2019edge}.  Nazeri et al. \cite{nazeri2019edgeconnect} proposed an edge generator to hallucinate edges in the edges of the missing regions, which can be regarded as an edge completion problem. Using edge images as structural guidance, high inpainting performance is achieved even for some highly structured scenes. 

Let $I_{gt}$ be the ground-truth image. $E_{gt}$ and $I_{gray}$ denote the edge map and grayscale image, respectively. In the edge generator, we use the masked grayscale image ${\widetilde{I}}_{gray}=I_{gray}\odot(1-M)$ and edge map ${\widetilde{E}}_{gt}=E_{gt}\odot(1-M)$ as the input. The image mask \emph{M} is defined by the binary image(1 for the missing region and 0 for the background). Here, $\odot$ denotes the Hadamard product. The generator predicts the edge map for the masked region.
\begin{equation}
 E_{pred}=G_e (\tilde{I}_{gray},\tilde{E}_{gt}, M).
  \label{equ:dt}
\end{equation}
Our edge discriminator $D_e$ receives  $E_{gt}$ and $E_{pred}$ conditioned on $I_{gray\ }$\ as inputs and predicts whether the edge map is real or fake. The edge generator is trained with an objective comprising of the hinge variant of GAN loss \cite{miyato2018spectral} and feature-matching loss \cite{wang2018high}. The hinge loss is effectively used in binary classification \cite{bartlett2008classification}. We believed that the hinge loss we used in this task would be effective because we train the edge generator using binary edge maps.
\begin{equation}
 \mathcal{L}_{G_e} = \lambda_{hinge}\mathcal{L}_{hinge}+\lambda_{FM}\mathcal{L}_{FM},
  \label{equ:dt}
\end{equation}
where $\lambda_{hinge}$ and $\lambda_{FM}$ are regularization parameters. The generator and discriminator with hinge loss are defined as follows: 
\begin{equation}
\begin{split}
 \mathcal{L}_{hinge}=-\mathbb{E}_{I_{gray}}[D_e(E_{pred},I_{gray})],
  \label{equ:dt}
  \end{split}
\end{equation}
\begin{equation}
\begin{split}
 \mathcal{L}_{D_e}=\mathbb{E}_{\left(E_{gt},I_{gray}\right)}\left[max(0, 1-{D_e}\left(E_{gt},I_{gray}\right)\right] \\ + \mathbb{E}_{I_{gray}}[max(0, 1+D_e(E_{pred},I_{gray})].
  \label{equ:dt}
  \end{split}
\end{equation}
The feature-matching loss, $\mathcal{L}_{FM}$, compares the activation maps in the intermediate layers of the discriminator. This stabilizes the training process by forcing the discriminator to produce an output that is similar to the real image. This is similar to perceptual loss \cite{gatys2016image,gatys2015texture} in which activation maps are compared with the feature maps of the pre-trained VGG network  \cite{simonyan2014very}. The feature-matching loss $\mathcal{L}_{FM}$ is defined as
\begin{equation}
 \mathcal{L}_{FM}{=}\mathbb{E}\bigg[\sum_i^L\frac{1}{N_i}||D_e^{(i)}(E_{gt})-D_e^{(i)}(E_{pred})||_1\bigg],
  \label{equ:dt}
\end{equation}
where $N_i$ is the number of elements and $D_e$ is the activation in the \emph{i}th layer of the discriminator. 

\subsection{Image Completion Network} 
After obtaining $E_{pred}$, $G_c$ generates a complete colored image. The masked color image ${\widetilde{I}}_{gt}=I_{gt}\odot(1-M)$ and conditional composite edge map $E_{comp}=E_{gt}\odot\left(1-M\right)+E_{pred}\odot$ \emph{M} were used as inputs. 
\begin{equation}
I_{pred}=G_c\left({\widetilde{I}}_{gt}, E_{comp}\right),
  \label{equ:dt}
\end{equation}
where $I_{pred}$ denotes the final output result. This network is trained over a joint loss that comprises $\ell_1$ loss, adversarial loss, perceptual loss, and style loss. To ensure proper scaling, the $\ell_1$ loss is normalized by the mask size. We employ adversarial loss in our generator to generate realistic results.
\begin{equation}
\begin{split}
 \mathcal{L}_{adv}{=}\mathbb{E}_{\left(I_{gt},E_{comp}\right)}\left[\log{D_c}({I_{gt},E_{comp}})\right] \\ + \mathbb{E}_{E_{comp}}[log[1-D_c(I_{pred},E_{comp})]].
  \label{equ:dt}
  \end{split}
\end{equation}
Perceptual loss is defined as follows:
\begin{equation}
  \mathcal{L}_{perc}{=}\mathbb{E}\bigg[\sum_i\frac{1}{N_i}||\varphi_i(I_{gt})-\varphi_i(I_{pred})||_1\bigg].
  \label{equ:dt}
\end{equation}
$\mathcal{L}_{perc}$ penalizes results that are not perceptually similar to features by defining a distance measure between activation maps of a pre-trained network. The style loss is defined as follows:
\begin{equation}
 \mathcal{L}_{style}{=}\mathbb{E}_j\bigg[||G_j^{\varphi}(\widetilde{I}_{pred})-G_j^{\varphi}(\widetilde{I}_{gt})||_1\bigg].
  \label{equ:dt}
\end{equation}
We choose to use style loss by Sajjadi et al. \cite{sajjadi2017enhancenet} to be an effective tool to combat “checkerboard” artifacts caused by the transpose convolution layers \cite{odena2016deconvolution}.  The final total loss is defined as
\begin{equation}
 \mathcal{L}_{G_c}{=}\lambda_{\ell_1}\mathcal{L}_{\ell_1}+\lambda_{adv}\mathcal{L}_{adv}+\lambda_p\mathcal{L}_{perc}+\lambda_{style}\mathcal{L}_{style}.
  \label{equ:dt}
\end{equation}
To address color tone mismatch problems in the previous outpainting methods, we use a large proportion for style loss. For our experiments, we set $\lambda_{\ell_1}$ as 1,\ $\lambda_{adv}$ as 0.2,\ $\lambda_p$ as 0.1, and $\lambda_{style}$ as 250.
%------------------------------------------------------------------------- 
\begin{figure}[b!]
\centering
\includegraphics[width=80mm,height=4cm]{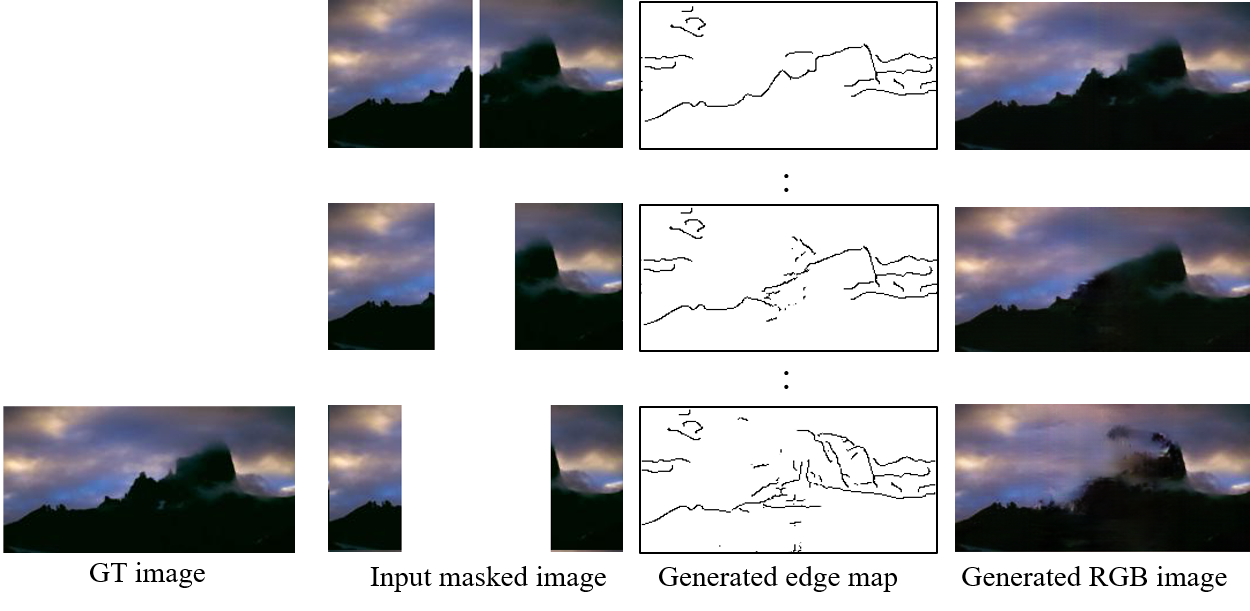} 
\caption{
The process of image completion based on the edge map progressively generated by the edge map generator.}
\label{fig:9}
\end{figure}

\begin{table*}[t!]
\centering
% \begin{tabular}{C{23mm}C{5mm}C{5mm}C{5mm}C{5mm}C{1mm}}
% \begin{tabular}{p{23mm}p{4mm}cp{5mm}Ccp{5mm}}

\begin{tabular}{cccccc}
\hline

Method         &  IS            & FID            & PSNR           & SSIM & BRISQUE        \\ \hline\hline
Pix2Pix \cite{isola2017image}        & \textbf{2.82}          & \textbf{19.73}          &    \textbf{-}          & \textbf{-}  &  \textbf{-}            \\ \hline
GLC\cite{iizuka2017globally}            & \textbf{2.81}          & \textcolor{blue}{\textbf{14.82}}          &  \centering  \textbf{-}          &  \textbf{-}   &  \textbf{-}            \\ \hline
CA\cite{yu2018generative}             & \textbf{2.93}          & \textbf{19.04}          & \textbf{20.42}          & \textbf{0.84} & \textbf{24.46}          \\ \hline
StructureFlow\cite{ren2019structureflow}  & \textcolor{blue}{\textbf{3.10}}          & \textbf{15.69}          & \textcolor{red}{\textbf{22.94}} & \textcolor{blue}{\textbf{0.85}} & \textbf{26.36}          \\ \hline
NS-OUT\cite{yang2019very}         & \textbf{2.85}          & \textcolor{red}{\textbf{13.71}} & \textbf{19.53}          & \textbf{0.72} & \textbf{\textcolor{blue}{23.59}}          \\ \hline
Proposed(w/o BR)   & \textbf{3.07}          & \textbf{17.75}          & \textbf{21.41}          & \textbf{0.84} & \textbf{23.62}          \\ \hline
    Proposed (with BR) & \textcolor{red}{\textbf{3.20}} & \textbf{15.72}         & \textcolor{blue}{\textbf{22.45}}          & \textcolor{red}{\textbf{0.86}} & \textcolor{red}{\textbf{21.61}} \\ \hline
\end{tabular}

\vspace*{2mm}
\caption{Quantitative results for conventional and proposed models on  SUN dataset \cite{xiao2010sun}. Evaluation of Inception Score (IS)  \cite{salimans2016improved}, PSNR and SSIM \cite{wang2004image} (the higher, the better), \'Frenchet Inception Distance FID  \cite{heusel2017gans} and BRISQUE \cite{mittal2012no} (the lower, the better). Images from the validation set had an IS of 3.479 (The best result of each column is red, the second-best result is blue, and BR means Bidirectional Rearrangement).}
\label{table1}
\end{table*}

\subsection{Horizontally progressive step Learning Method }
GANs are difficult to train because of problems such as mode collapse, non-convergence, and vanishing gradient \cite{arjovsky2017wasserstein}. These problems also apply to the image outpainting task where more areas than expected are generated without context. We propose a simple but effective training technique to stabilize GAN training from the perspective of image outpainting by horizontally increasing the mask size (Fig. ~\ref{fig:9}.). Generators are usually successful in a small mask task but have difficulty in a wide mask task \cite{abbas2019learning}. Therefore, we divide the learning step by mask size so that the model can learn more stably while increasing the mask size. The mask is divided into 32 steps and gradually grows at each step. The initial mask size is 3.125\% of the input image and linearly increases, and the final mask size is set at 50\% of the input image.

\section{Training}

When we training, we give the mask inside the images to proceed in the horizontally progressive step learning method. The training process corresponds to the gray box of Fig. ~\ref{fig:2}. When testing, the masked outside of the images is given as input and generates the outpainting images using the rearrangement method described above. 

\subsection{Training Setup }
Our model is implemented using the PyTorch framework. The network is trained with the SUN dataset \cite{xiao2010sun} comprising 256$\times$128 pixels. Considering our GPU memory, the batch sizes of $G_e$ and $G_c$ networks are 8 and 16, respectively. We use the AdamP optimizer \cite{heo2020slowing} with $\beta_1$ = 0 and $\beta_2$ = 0.99. Generators $G_e$ and $G_c$, are trained with the learning rate of $10^{-4}$ until the losses plateau separately. We lower the learning rate to $10^{-5}$ and continue to train $G_e$ and $G_c$ until convergence.

We use a Canny edge detector \cite{canny1986computational} to generate an edge map. The sensitivity of the Canny edge detector is controlled by the standard deviation of the Gaussian smoothing filter ($\gamma$). In the experimental results, it was the best when the $\gamma$ is 2.
Fig.~\ref{fig:9} illustrates the process of image completion based on the edge map generated by the edge map generator in each step.
% The first row indicates when the size of the mask corresponds to task 1; the second row shows the result of the intermediate task (16th mask), and the third row shows the result of the last task (32nd mask). 
The result of the image completion generator depends on the edge map.  

%------------------------------------------------------------------------- 
\section{Experiment}

\subsection{Quantitative Result}
Evaluating the quality of generated images in image outpainting has the same difficulty as that in GAN, as there are few restrictions on the created images. Note that the well-outpainted image has only to be photo-realistic while sharing the context naturally with the input image. Hence, we include non reference image quality metrics to evaluate the generate images. We used structural similarity index (SSIM) \cite{wang2004image}, peak signal-to-noise ratio (PSNR), Inception Score (IS)  \cite{salimans2016improved}, \'Frenchet Inception Distance (FID)  \cite{heusel2017gans}, and Blind/Referenceless Image Spatial Quality Evaluator (BRISQUE) \cite{mittal2012no} to measure the resulting quality with the state-of-the-art outpainting and inpainting algorithm. FID uses a pre-trained Inception-V3 model to measure the Wasserstein-2 distance between the actual image and the representation of the shape space of the painted image \cite{szegedy2016rethinking}. We evaluated the task of generating 50\% of the input image on the SUN dataset \cite{xiao2010sun} with other methods. The results for the SUN dataset are shown in Table.~\ref{table1}. Our model produced more natural images than other methods, as can be seen in the qualitative section. We further evaluated using the BRISQUE method \cite{mittal2012no} to evaluate how the naturalness of the image produced. A quantitative comparison shows that our model is superior to the other methods in all aspects except for the PSNR and the FID score. We believe that the combination of the listed metrics provide a better result in outpainting performance.

We additionally experimented with quantitative comparisons on the beach dataset \cite{sabini2018painting} for a more objective comparison of our model with other conventional outpainting methods. We tested our model using the pre-trained weights with the sun dataset \cite{xiao2010sun} without additional training on the beach dataset. We used the quantitative results from the SieNet \cite{zhang2020sienet} and compared them. The results for the beach dataset \cite{sabini2018painting} are listed in Table.~\ref{tab:beach}, and it can be seen that our model has better structural strength compared to other models.

\subsection{Qualitative Result}
\begin{figure*}[t!]
% \centering
\includegraphics[width=\textwidth]{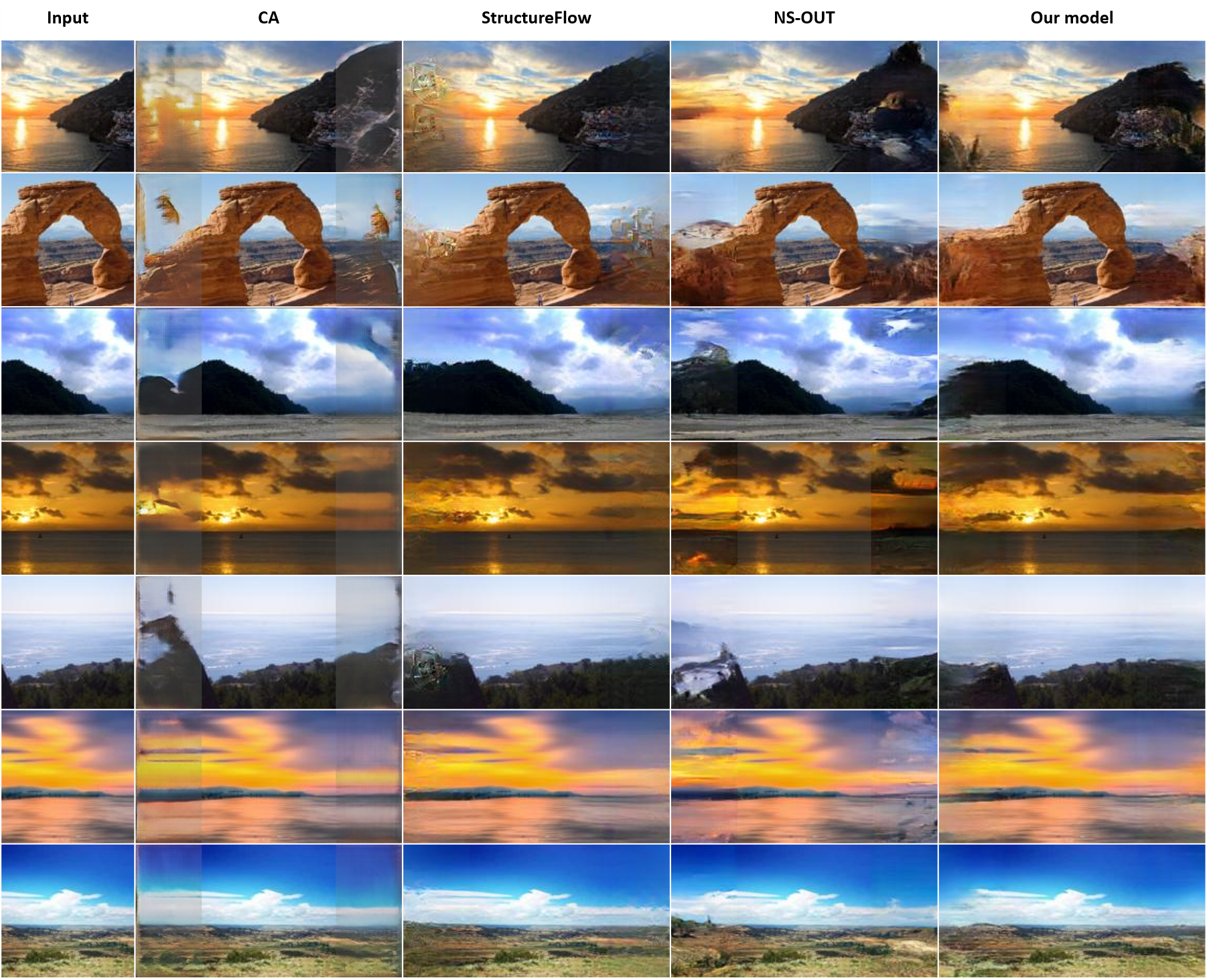}
\caption{
Qualitative results for conventional and proposed models on SUN dataset \cite{xiao2010sun}: CA\cite{yu2018generative}, StructureFlow\cite{ren2019structureflow}, NS-OUT \cite{yang2019very}, and our model.} 
\label{fig:5}
\end{figure*}

Fig.~\ref{fig:5} shows a qualitative result comparison image between the proposed method and the state-of-the-art methods NS-OUT \cite{yang2019very}, and the inpainting methods CA  \cite{yu2018generative} and StructureFlow \cite{ren2019structureflow}.
CA \cite{yu2018generative} produced images that show inconsistent objects near the missing image parts and show large color differences from the original image. The outpainting algorithm NS-OUT \cite{yang2019very} and the inpainting algorithm StructureFlow \cite{ren2019structureflow} produced images that are more clearer, but image distortion and blurring in the generated image still exist.  Our proposed method produced images that are clearly connected with the original image and are smoother and more consistent than those obtained by NS-OUT and StructureFlow.

\begin{table}[t!]
\centering
\begin{tabular}{cccc}
\hline
Method            & SSIM            & PSNR             &  \\ \hline\hline
Image-Outpainting\cite{sabini2018painting} & \textbf{0.338}          & \textbf{14.625}          &  \\ \hline
Outpainting-srn\cite{wang2019wide}   & \textbf{0.513}          & \textbf{18.221} &  \\ \hline
SieNet\cite{zhang2020sienet}            & \textbf{\textcolor{blue}{0.646}}          & \textbf{\textcolor{red}{20.796}} &  \\ \hline
Proposed              & \textbf{\textcolor{red}{0.810}} & \textbf{\textcolor{blue}{18.957}}          &  \\ \hline
\end{tabular}
\vspace*{2mm}
\caption{Performance on beach dataset \cite{sabini2018painting} (The best result of each column is red and the second-best result is blue.).}
\label{tab:beach}
\end{table}

\subsection{Ablation}

We conducted ablation studies to demonstrate the necessity of introducing bidirectional boundary rearrangement. A comparison of the quantitative results of our architecture and the model without bidirectional rearrangement (w/o BR) is shown in Table.~\ref{table1}. According to the results, the model with bidirectional rearrangement method (with BR) effectively improved the performance at all values.
% \subsection{Hinge Loss}

\begin{figure}[t!]
\centering
\includegraphics[width=82mm]{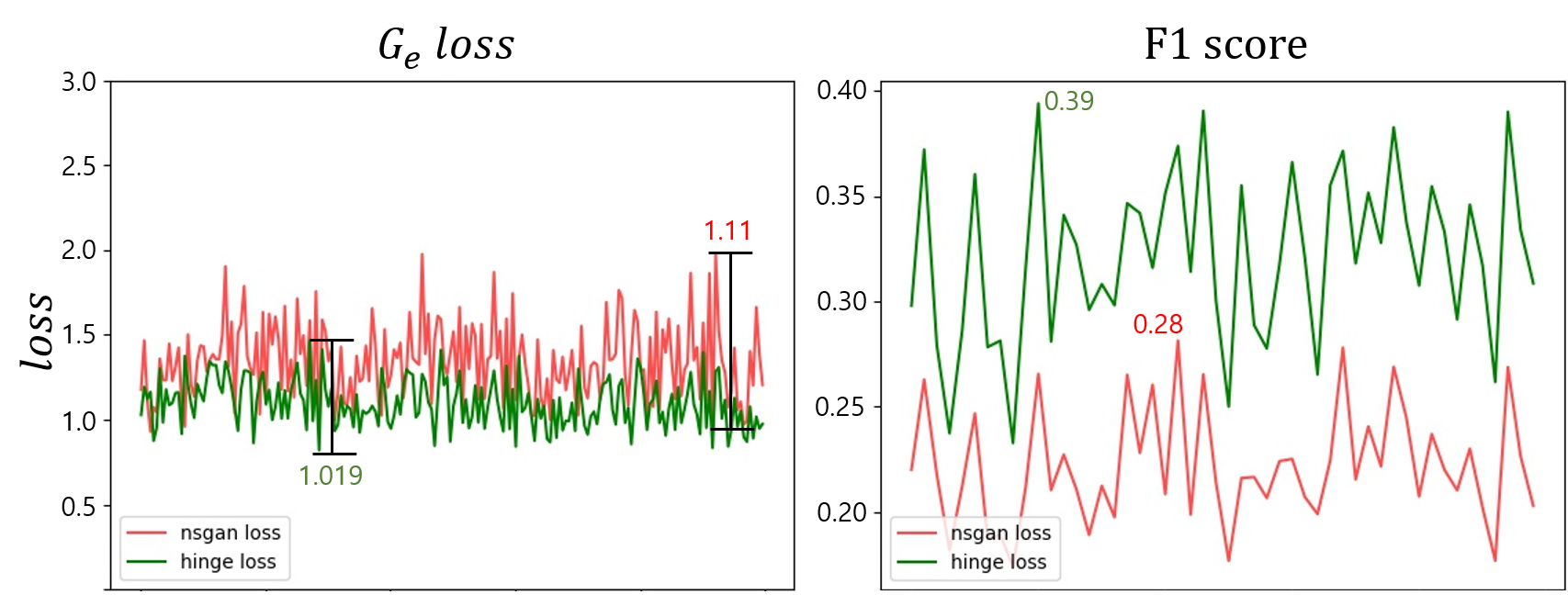} 
\caption{
Loss value and F1 score of the edge map generator in the last training step. In the left image, the peak-to-peak amplitude of the hinge loss is 1.019 and the nsgan loss is 1.11. It can be seen that the hinge loss converges to a smaller amplitude. In the right image, the highest F1 score of hinge loss is 0.39 and nsgan loss is 0.28.}
\label{fig:loss}
\end{figure}

As shown in Fig.~\ref{fig:loss}, hinge loss is more effective than Non-Saturating GAN loss (nsgan loss) \cite{goodfellow2014generative}. In the case of using the hinge loss function, the fluctuation of the loss vibrates to a smaller value during the learning process, and the learning process is more stable. we observed a higer F1 score than nsgan loss. The edge guided method useful not only for image outpainting, but also for image inpainting \cite{nazeri2019edgeconnect}, super-resolution  \cite{nazeri2019edge}, and various other fields \cite{chai2020mri,jiang2019edge}. Therefore, it is believed that the hinge loss in other fields as well as binary data such as edge maps will help improve the performance. 

\begin{figure}[b!]
\centering
\includegraphics[width=80mm]{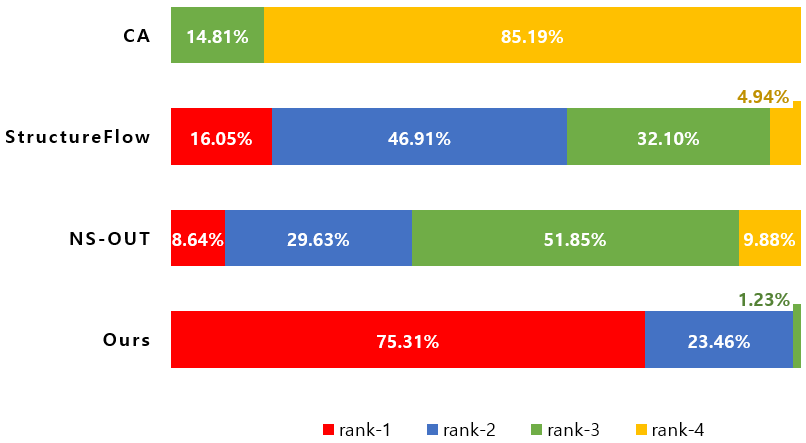} 
\caption{User study results between our proposed method and conventional methods (CA \cite{yu2018generative}, StructureFlow \cite{ren2019structureflow}, NS-OUT \cite{yang2019very}). Our method obtained the most superior results.
}
\label{fig:user}
\end{figure}

\subsection{User Study}

We performed a user study to compare the image outpainting performance of our proposed method with other image inpainting/outpainting methods, such as CA \cite{yu2018generative}, StructureFlow \cite{ren2019structureflow} and NS-OUT \cite{yang2019very}, using benchmark datasets. A total of 27 experts in image processing participated in anonymous voting to evaluate naturalness from randomly selected 30 images to ranks 1 to 4. Fig.~\ref{fig:user} shows the summarized results. Each rank has a total of 810 votes, where our proposed method obtained 610 votes and 190 votes in rank 1 and rank 2, respectively. Consequently, the comparison result verified that our proposed method outperforms other methods in generating a more visually clear image to human raters.

\section{Limitation}
The proposed method produced a natural image well when the information at both ends of the image is similar. However, when the assumption, where information on both ends of an image is similar, is broken, our method generated an output image with different characteristics with the ground truth, but produced natural images. As can be seen from the quantitative and qualitative results above, at the sun and beach datasets showed overall good results. We expect results to be poor for datasets that are more complex and have little similar information at both ends. We will try to consider theses factors in the future work.

\section{Conclusion and Future Work}
In this paper, we proposed a novel image outpainting method composed of three approaches. First, we rearranged the bidirectional boundary regions to address the lack of information when filling the image outward. The previous methods had difficulties in generating images because the adjacent information was not sufficient to generate large empty areas. However, the edges from our structural edge map generator worked as a guideline to reflect semantic information from the given regions to the unknown ones. In addition, with the rearrangement of the boundary regions, increased adjacent information prevented the large unknown areas from being filled with repetitive pixel values. Second, we present a hinge loss specialized in generating binary images such as edge maps. Third, the training step was divided according to different horizontal mask sizes so that the model was trained stably as well as naturally generated images outside and inside. Through multiple steps, we can obtain a wider and better-quality image. We evaluated our model compared to conventional image inpainting and outpainting methods in terms of qualitative and quantitative measurements. The experimental results show that our method outperformed the other methods.

In future work, we will explore how to generate outpainting images on horizontal and vertical directions with the same model simultaneously. Besides, we plan to design a novel model whether to selectively refer to different information on both ends so that a natural image can be robustly generated even in the case of images with different information on both ends.

\section{Acknowledgments}
This research was supported by the MSIT(Ministry of Science and ICT), Korea, under the ITRC(Information Technology Research Center) support program(IITP-2020-2018-0-01421) supervised by the IITP(Institute for Information \& Communications Technology Planning \& Evaluation), the National Research Foundation of Korea(NRF) grant funded by the Korea government(MSIT)(No. 2020M3H4A1A02084899) and the National Research Foundation of Korea (NRF) grant funded by the Korea government (MSIT) (No. 2018R1D1A1B07048421)  
%-------------------------------------------------------------------------

{\small
\bibliographystyle{ieee_fullname}
\bibliography{egbib}
}

\end{document}